\documentclass{vgtc}                          % final (conference style)
\graphicspath{{figures/}{pictures/}{images/}{./}} % where to search for the images

\usepackage{times}                     % we use Times as the main font
\usepackage{tabu}                      % only used for the table example
\usepackage{booktabs}                  % only used for the table example
\usepackage{lipsum}                    % used to generate placeholder text
\usepackage{mwe}                       % used to generate placeholder figures
\usepackage{amsmath}
\usepackage{subcaption}
\usepackage{tabularx}
\usepackage{array}
\usepackage{float}

\usepackage{mathptmx}                  % use matching math font
\usepackage{bbm}

\onlineid{0}

\vgtccategory{Research}

\vgtcinsertpkg

\title{Spatial Action Review: A Visual Analytics Dashboard for Auditing Language-to-Action Hand-offs in Electron Microscopy}

\author{Samia Mohinta\thanks{Correspondence to e-mail: sm2667@cam.ac.uk}\\ %
        \scriptsize University of Cambridge, UK %
\and Albert Cardona\\ %
     \scriptsize MRC Laboratory of Molecular Biology, UK} %
\teaser{
\centering
  \includegraphics[width=\textwidth]{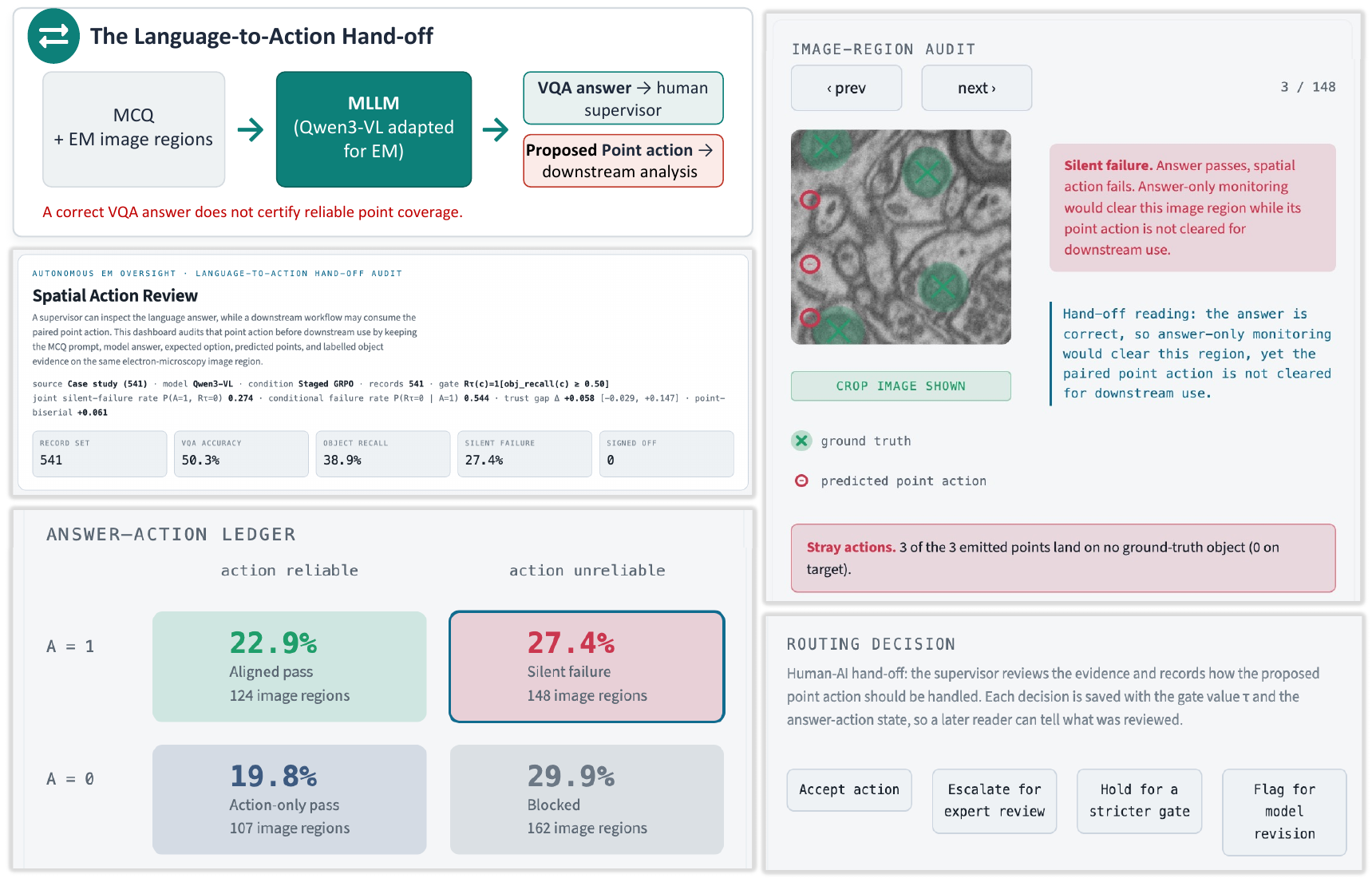}
\caption{
\textbf{\emph{Spatial Action Review} dashboard auditing the language-to-action hand-off for the EM-adapted
Qwen3-VL case study.} A single MLLM answers a multiple-choice question about an EM image region and
proposes a point action for the same region (top left). The summary panel and the answer-action
ledger report the run at the action-reliability gate in force (left). The image-region audit shows one silent failure,
where the answer is correct while all three emitted points fail to land on any annotated mitochondrion (top
right), and the routing panel records a human supervisor's disposition (bottom right). Dashboard is live at \href{https://spatialactionreview.streamlit.app/}{this URL}.}
  \label{fig:teaser}
}

\abstract{
Multimodal large language models (MLLMs) are increasingly explored as interfaces for scientific
image analysis, where a visual question-answering (VQA) response may be paired with a spatial
output that guides a downstream stage. In practice, a supervisor reads the language answer, while a
downstream workflow such as segmentation, reacquisition, or region review consumes the point-set
output. We call this transition, from inspecting the answer to relying on its paired point action,
the \emph{language-to-action hand-off}. A \emph{silent failure} occurs when the answer is correct
while the paired action misses the annotated objects the downstream stage needs, so answer-based
oversight clears a region whose action is unreliable. We introduce \emph{Spatial Action Review}, an
interactive visual analytics dashboard for auditing this failure mode in electron microscopy (EM)
mitochondria analysis. It links paired answer-action records through an answer-action ledger, a
task-by-dataset risk map, and an image-region audit view, so that linked selections connect
aggregate patterns to image evidence, while an adjustable action-reliability gate supports
re-audit. The review ends in a \emph{human-AI hand-off}, where a supervisor records whether the
action is accepted, escalated for expert review, held under a stricter gate, or flagged for model
revision. Across 541 image regions from an EM-adapted Qwen3-VL case-study run, our dashboard shows that point actions fail
the gate in 54.4\% of records with a correct VQA response, and 27.4\% of all records are silent
failures. A correct answer is associated with only a 5.8-percentage-point higher probability of a
reliable action, a trust gap whose bootstrap interval spans zero, and the point-biserial
correlation between answer correctness and object coverage is 0.061. This weak coupling persists
when the same measures are recomputed across five model conditions on 753 matched image regions.
\emph{Spatial Action Review} makes answer-action mismatches visible and ties them to image evidence
and a recorded decision, before MLLM outputs enter increasingly autonomous scientific workflows.
}


\begin{document}

% \begin{figure}[!ht]
%   \centering
%   \includegraphics[width=\linewidth]{figures/fig1_illustration.pdf}
%   \caption{
%   \textbf{Silent failure in a language-to-action hand-off for electron microscopy analysis.}
%   An MLLM provides an acceptable language answer while producing point actions that miss the relevant mitochondria. Green crosses indicate ground-truth target locations and red circles indicate model-predicted points. The example motivates auditing spatial actions directly before they are used by a closed-loop image analysis workflow.
%   }
%   \label{fig:silent_failure}
% \end{figure}


\firstsection{Introduction}
\maketitle
\label{sec:introduction}

Autonomous laboratories are moving from scripted automation toward systems that plan, observe,
and act with increasing independence~\cite{fieldhouse2026robots,zheng2025automation}.
Multimodal large language models (MLLMs) are attractive components in such systems because they
connect visual observations with natural-language instructions and
responses~\cite{zhang2024bioimage}, and recent models can also emit spatial coordinates on
request~\cite{peng2023kosmos2,qwen3vl}. A single model therefore produces two outputs of different
kinds. One is a visual question-answering (VQA) response that a scientist reads. The other is a spatial output that a
downstream stage consumes, for example to seed a segmentation from point
prompts~\cite{kirillov2023sam}, prioritize an image region for review, or select a candidate field
of view for reacquisition. We call a model-generated spatial output intended for such downstream
use a \emph{spatial action}.

These two outputs do not, however, receive the same scrutiny. A language answer is presented to a
scientist who can inspect and question it, and in an annotated evaluation it can also be checked
against an expected answer. A spatial action, by contrast, may reach a downstream stage without an
equivalent point of inspection. We call this operational transition, from inspecting the language
answer to relying on its paired spatial action, the \emph{language-to-action hand-off}. Both
outputs are generated by the same model for the same image region, so we treat them as a single
paired record rather than as two independent scores. A \emph{visual oversight gap} then arises
whenever scrutiny stops at a readable answer while a downstream stage depends on its paired
spatial action. This hand-off is distinct from the \emph{human-AI hand-off}, in which a human
supervisor, typically a domain scientist, decides how a proposed action should be handled after
reviewing available evidence. We audit the first hand-off and end in the second.

\Cref{fig:teaser} (top right) shows what the visual oversight gap costs in our setting, where an EM-adapted MLLM
answers a multiple-choice question about an image region and, in a separate request, returns one
predicted point per visible mitochondrion. For this region the answer is correct, while all three
emitted points land off every annotated mitochondrion. A supervisor reading the answer sees a
region that passed, yet the output that leaves the region for downstream use is unreliable. We call
this a \emph{silent failure}. The answer is correct, but the paired action fails a predefined
action-reliability criterion, which we call the \emph{action-reliability gate}. An action passes
the gate when its predicted points collectively cover at least half of the annotated objects in
the region (\Cref{sec:methodology}). The failure is silent with respect to answer-based oversight
because the output a supervisor inspects appears correct. Because a language answer is more
readily inspected, a correct answer can be treated as clearance for its paired spatial action, and
that clearance can be misplaced. If the action proceeds, it may seed a segmenter at the wrong
location or select an unsuitable field of view. The oversight question is therefore not only
whether the model answered correctly, but also whether spatial evidence supports allowing the
paired action to proceed.

Evaluations that report answer correctness alone cannot expose this failure mode. Microscopy
vision-language benchmarks assess expert question answering and broad microscopy
understanding~\cite{lozano2024mubench,burgess2025microvqa}, biomedical benchmarks add grounded
classification and detection as separate task configurations~\cite{dcunha2026mmbu}, and recent
general-domain benchmarks score language answers and visual evidence
jointly~\cite{nasrazadani2026vistaqa,wang2026evqa}. These advances make answer-grounding coupling
measurable, but a joint score does not preserve answer-action states, show where silent failures
concentrate across tasks and datasets, or support a supervisor's decision before a proposed action
is used. Our question is therefore not whether an answer and a spatial output can be scored
together, but how their relationship can be inspected.

We instantiate this oversight problem in an MLLM-driven workflow for automated electron microscopy
(EM) analysis of mitochondria, using Qwen3-VL~\cite{qwen3vl} as the model under audit
(\Cref{sec:methodology}). EM is a suitable testbed for two reasons. First, many analysis steps
depend on local object evidence rather than on a global image
label~\cite{lucchi2013,wei2020mitoem}, so a misplaced point can misdirect the next step.
Second, ground-truth annotations allow both outputs to be scored on the same image region. The
oversight problem itself, however, belongs to any workflow that relies on paired language and
spatial outputs rather than to one particular model or dataset.

To close this gap before an action is used, we introduce \emph{Spatial Action Review}, an
interactive visual analytics dashboard that audits the relationship between the two outputs in
each paired record. It resolves records into four answer-action states, localizes silent failures
across task and dataset, links states and failures to image-region evidence, supports re-audit
under an adjusted gate, and ends in a recorded supervisory decision (\Cref{sec:methodology}).
Across 541 image regions from an audited case study, the dashboard reveals that 54.4\% of point
actions paired with a correct answer fail the gate, that silent failures account for 27.4\% of all
records, and that a correct answer is associated with only a 5.8-percentage-point higher
probability of a reliable action, with a bootstrap interval spanning zero. A separate matched audit likewise finds weak coupling across five model conditions on 753 image
regions (\Cref{sec:results}). Answer correctness is therefore a weak clearance signal in this case-study
run, which is precisely the oversight condition the dashboard is designed to expose.

Our contributions are as follows.
\begin{itemize}
  \item We formulate the language-to-action hand-off as a visual oversight problem in closed-loop
    scientific image analysis and distinguish it from a human supervisor's subsequent decision
    about how a spatial action should be handled.
  \item We define paired answer-action records for EM image regions and report answer-conditioned
    reliability measures over them, namely silent-failure rate, trust gap, and point-biserial
    correlation.
  \item We present \emph{Spatial Action Review}, a dashboard providing a linked audit path from
    answer-action states and task-dataset risk to image-region evidence, with gate-dependent
    re-audit and a recorded supervisory routing decision.
\end{itemize}

\Cref{fig:teaser} gives an overview of the case study and the linked review views. The dashboard is live at
\href{https://spatialactionreview.streamlit.app/}{\texttt{spatialactionreview.streamlit.app}}
and the source code and audit records are at
\href{https://github.com/Mohinta2892/SpatialActionReview}{\texttt{github.com/Mohinta2892/SpatialActionReview}}.

\section{Related Work}
\label{sec:related_work}

\textbf{Visual analytics for inspection, reliance, and audit.} Visual analytics treats interactive
inspection and drill-down as central mechanisms for reasoning about complex computational
processes~\cite{heer2012interactive}, and systems built for machine learning support model
understanding, diagnosis, comparison, and
refinement~\cite{liu2017towards,hohman2018visual,choo2018visual}, including coordinated views at
instance and subset level~\cite{kahng2018activis} and interactive probing across examples and
subgroups~\cite{wexler2019whatif}. Their shared lesson is that aggregate metrics are insufficient
and that users need linked views connecting summary patterns to the individual evidence underlying
them. We apply this principle to the language-to-action hand-off, through linked views that carry a
supervisor from an aggregate state to the image region behind it.

Trust in automation depends on appropriate reliance, which requires knowing when an output should
be trusted in a particular case rather than only how accurate a system is on
average~\cite{lee2004trust,amershi2019guidelines,shneiderman2020human}. A silent failure creates
exactly this form of inappropriate reliance, because a readable answer invites confidence while
the spatial action intended for downstream use remains unreliable. Explainability methods expose
the evidence behind an individual prediction~\cite{ribeiro2016lime,selvaraju2017gradcam}, but not
whether the paired spatial action satisfies its own reliability criterion on the same image
region. That relationship is the transparency question we address.

Moreover, autonomous scientific settings add a decision point. Human-in-the-loop work in medical
image analysis argues that human involvement remains important during deployment rather than only
during annotation or training~\cite{budd2021survey}, and in automated EM such control has been
proposed for experiments in which a learning agent guides acquisition while a person steers
policy~\cite{kalinin2023human}. A decision of this kind is accountable only if it can be
reconstructed afterwards, which is the concern of audit-oriented work on traceable
records~\cite{raji2020closing} and of provenance research in
visualization~\cite{ragan2016provenance}. Our human-AI hand-off differs in scope. It does not
steer acquisition policy but adjudicates a single paired record after its image-level evidence has
been inspected. It nonetheless satisfies both conditions above, since a human makes the
deployment-time decision and the decision is stored with the gate value and answer-action state
under which it was made, so it can be reconstructed later.

\textbf{MLLMs for scientific image analysis.} MLLMs provide prompt-based interfaces for
image-conditioned tasks, from few-shot question answering and
description~\cite{alayrac2022flamingo,li2023blip2} to open-ended instruction
following~\cite{liu2023llava} and spatially explicit outputs such as boxes and point
coordinates~\cite{peng2023kosmos2}. In bioimage analysis this interface is attractive because
scientists can express intent in natural language rather than requiring a task-specific tool for
every query or modality~\cite{zhang2024bioimage}, which also makes MLLMs relevant to autonomous
laboratories~\cite{zheng2025automation,fieldhouse2026robots}. Adapting such a model to a
specialized domain is well studied, through supervised fine-tuning (SFT) on labeled
examples~\cite{ouyang2022training,liu2023llava}, reward-based methods such as Group Relative
Policy Optimization (GRPO) ~\cite{shao2024deepseekmath}, and parameter-efficient methods such as Low-Rank
Adaptation (LoRA)~\cite{hu2022lora}. In EM, such methods have been used to turn open-weight MLLMs into point-prompt generators for
downstream segmentation~\cite{mohinta2026adapting} and visual question answering with spatial
grounding~\cite{mohinta2026vqa}. We audit the model conditions this line of work produces; our
contribution is not a new adaptation method but the audit applied to their outputs
(see details in \Cref{sec:methodology}).

\textbf{Coupling language answers and spatial outputs.} Spatial grounding asks a model to localize
the object evidence a downstream stage requires, and has been studied through bounding
boxes~\cite{peng2023kosmos2,chen2023shikra}, free-form regions~\cite{you2023ferret}, and point
coordinates~\cite{cheng2025pointarena}. Microscopy evaluations, however, largely measure expert
question answering, whether perception and cognition across modalities and
organisms~\cite{lozano2024mubench} or research-level reasoning including hypothesis generation and
experiment proposal~\cite{burgess2025microvqa}, without pairing each answer with a spatial action
produced by the same model for the same image region. Biomedical and general benchmarks have begun
to close that gap to differing degrees. MMBU evaluates ungrounded classification, grounded
classification, and object detection across biomedical modalities~\cite{dcunha2026mmbu}, but as
distinct task configurations. Grounded classification supplies the region and asks for a label, so
the location is given rather than produced, while detection produces a location with no paired
language answer that could agree or disagree with it. VISTAQA requires both outputs on the same
sample, and its GROVE metric combines them multiplicatively through a per-sample geometric mean, so
a sample scores well only when the answer and the mask are both
correct~\cite{nasrazadani2026vistaqa}. Evidence-Backed Video Question Answering makes a comparable
demand for spatiotemporal evidence and reports substantial decoupling between answer accuracy and
visual perception~\cite{wang2026evqa}.

Our work shares this concern and differs in three ways. These benchmarks treat the spatial output
as evidence that justifies an answer, whereas here the point set is an action a downstream stage
consumes, which is what makes the transition between the two a hand-off rather than a scoring
problem. A joint score also collapses the four answer-action states, so it registers weak coupling without
separating a silent failure from a record where both outputs fail, and without showing where such
cases concentrate. Finally, our aim extends beyond measurement to a review path that
ends in a recorded decision. \emph{Spatial Action Review} therefore makes the language-to-action
hand-off inspectable rather than only measurable.

\section{Methodology}
\label{sec:methodology}

Auditing the language-to-action hand-off imposes three design requirements. The two outputs must
be scored for the same image region, so that their relationship can be inspected rather than
inferred from separate aggregates. That relationship must resolve into states a supervisor can
locate and trace to image-level evidence. The review must end in a decision recorded together with
the criterion under which it was made.

\emph{Spatial Action Review} implements these requirements as an interactive visual analytics
dashboard whose unit of analysis is the paired audit record, one language answer and the point
action produced for the same image region. A supervisor reads the summary state of the loaded
records, narrows to a task and dataset where the two disagree, opens the image-level evidence for
an individual record, and records how the proposed action should be handled. The first three steps
audit the language-to-action hand-off and the fourth is the human-AI hand-off. \Cref{fig:teaser}
shows this sequence for the case study, and \Cref{fig:interactions} displays the views that follow.

\subsection{Paired Answer-Action Records}
\label{sec:paired_records}

We instantiate the dashboard in automated EM mitochondria analysis. An image region is a
two-dimensional crop from an annotated EM volume, drawn from three public datasets that differ in
organism, tissue, and acquisition modality: Lucchi~\cite{lucchi2013,casser2020fast},
VNC~\cite{vnc2010}, and EM-H from MitoEM~\cite{wei2020mitoem}. Lucchi and EM-H provide instance
masks, so each mitochondrion is already one object, while for VNC objects are the connected
components of its binary masks, following Mohinta et al.~\cite{mohinta2026adapting,mohinta2026vqa}.
Every region therefore carries a set of annotated objects, its \emph{ground-truth object evidence},
from which the expected answers are derived and against which point actions are scored.

Each region yields one paired record with two channels. The answer channel is a multiple-choice
question with a label-derived expected answer, covering presence, counting, location, relation,
marked-region, and attribute judgments about mitochondria~\cite{mohinta2026vqa}. Marked-region
questions concern a highlighted part of the region and are grouped with location, giving five
displayed task types. The action channel asks the model to identify every mitochondrion visible in
the region and return one normalized point for each.

Both outputs come from the same model, prompted separately for the same region. Their outcomes can
still diverge, and detecting that divergence, a correct answer beside an unreliable action, is what
the audit is for. The pairing holds even when a question concerns a single property or object,
because the point request still targets the mitochondria visible in the region.

Each record holds the image-region identifier, dataset, task type, both prompts and their outputs,
the expected answer, the ground-truth object evidence, the metrics of \Cref{sec:scoring}, and the
resulting action verdict, serialized as one \texttt{inspector\_data.json} record. Every dashboard
value derives from these records, so each aggregate traces to the evidence behind it. The full
schema, the crop convention, and the task and answer-option distributions are given in the
Supplementary Material.

\subsection{Audit Sources and Model Conditions}
\label{sec:setup}

The dashboard inspects any collection of records in this schema and exposes three audit sources. The
\emph{case study} holds 541 records from Qwen3-VL~\cite{qwen3vl} adapted with \emph{Staged GRPO} and
is traced end to end in \Cref{sec:results}. The \emph{SFT-variant audit} holds a matched set of 753
image regions scored under five model conditions obtained from related EM
work~\cite{mohinta2026vqa}: \emph{Zero-shot}, the base instruction-tuned model with no task-specific
training, and four supervised variants, namely \emph{Perception SFT} on answer examples,
\emph{Grounding SFT} on point-action examples, \emph{Staged SFT} applying answer then grounding
supervision in sequence, and \emph{Joint SFT} mixing both. \emph{Staged GRPO} continues from the
Staged SFT adapter with a reinforcement-learning objective on the action channel~\cite{mohinta2026adapting, mohinta2026vqa,shao2024deepseekmath}.
The conditions therefore differ in how the two channels are adapted, while all supervision derives
from the same object-level ground truth that defines the expected answers and the object evidence.
We audit these outputs and do not contribute the adaptation; the case study and the SFT-variant
audit are reported separately, since they hold different model conditions.

A third source is user-provided. The dashboard audits any run exported in the
\texttt{inspector\_data.json} schema, with an optional zip of crop images, so a user can upload
their own records; a preparation script and the schema are documented with the source.

\subsection{Scoring Answer Correctness and Action Reliability}
\label{sec:scoring}

Let \(c\) denote an image region. Answer correctness is
\begin{equation}
  A(c)=\mathbbm{1}[\hat{y}(c)=y(c)],
  \label{eq:answer_correctness}
\end{equation}
where \(\hat{y}(c)\) is the generated option letter or normalized answer text and \(y(c)\) the
expected answer. Because the number of options varies across questions (from \(K=1\) to \(K=9\); see
Supplementary Material), the dashboard reports the record-wise uniform-choice baseline, the mean of
\(1/K\), beside answer accuracy.

Let \(G(c)\) be the ground-truth object evidence and \(P(c)\) the predicted point set. An annotated
object is covered when at least one predicted point falls inside its mask, and object recall is the
covered fraction,
\begin{equation}
  \mathrm{obj\_recall}(c)=\frac{|\{g\in G(c):g\text{ is covered}\}|}{|G(c)|}.
  \label{eq:object_recall}
\end{equation}
The supplementary material gives the convention for image regions containing no annotated
mitochondria.

Object recall is reduced to a binary action verdict,
\begin{equation}
  R_\tau(c)=\mathbbm{1}[\mathrm{obj\_recall}(c)\geq\tau],
  \label{eq:action_reliability}
\end{equation}
where \(\tau\) is the action-reliability gate. We report \(\tau=0.5\), which passes a region whose
points cover at least half of its annotated mitochondria. This is a permissive audit criterion
rather than a validated safety threshold, and the supervisor can adjust it (\Cref{sec:interaction}).
A silent failure is a record with \(A(c)=1\) and \(R_\tau(c)=0\).

Object recall measures coverage and does not penalize points that fall outside every annotated
object, so a region can pass the gate despite such errors. The image-region audit view therefore
also reports point-in-mask precision, the fraction of emitted points that fall inside an annotated
object and the precision counterpart of \Cref{eq:object_recall} under the same mask-membership rule,
together with a coordinate-derived stray-point count for predictions unmatched to object centroids (see Supplementary Material). The gate
narrows what requires inspection without replacing it.

\subsection{Answer-Conditioned Reliability}
\label{sec:answer_conditioned}

Answer correctness and the action verdict define four answer-action states,
\begin{equation}
\begin{aligned}
\text{Aligned pass}     &: A(c)=1,\;R_\tau(c)=1,\\
\text{Silent failure}   &: A(c)=1,\;R_\tau(c)=0,\\
\text{Action-only pass} &: A(c)=0,\;R_\tau(c)=1,\\
\text{Blocked}          &: A(c)=0,\;R_\tau(c)=0.
\end{aligned}
\label{eq:behavioral_states}
\end{equation}
Aligned pass and silent failure partition the answer-correct records, so their rates sum to answer
accuracy at every gate.

The joint rate \(P(A=1,R_\tau=0)\) gives the prevalence of silent failures across the active source,
while the answer-conditioned rate \(P(R_\tau=0\mid A=1)\) gives how often the action fails among
records whose answer is correct. The risk map decomposes the joint rate by task \(t\) and dataset
\(d\),
\begin{equation}
  S(t,d)=P\!\left(A=1,\,R_\tau=0\mid T=t,\,D=d\right).
  \label{eq:pairwise_risk}
\end{equation}
Its denominator is every record in the pair, so the interface reports the record count beside each
rate.

The trust gap
\begin{equation}
  \Delta=P(R_\tau=1\mid A=1)-P(R_\tau=1\mid A=0)
  \label{eq:trust_gap}
\end{equation}
measures whether action reliability differs with answer correctness, and a value near zero means
answer correctness separates reliable from unreliable actions weakly. We estimate its interval by
nonparametric percentile bootstrap over paired records, drawing \(B=2000\) resamples with
replacement using seed 0 within each condition. Because \(\Delta\) compares two binary variables, we
also report the point-biserial correlation~\cite{lev1949point} between binary answer correctness and
graded object recall. The states, both silent-failure rates, and \(\Delta\) all depend on \(\tau\),
whereas this correlation does not, so it serves as the gate-independent summary of answer-action
coupling.

\subsection{Linked Views and Interaction}
\label{sec:interaction}

A summary panel (\Cref{fig:teaser}, upper left) reports the active audit source, model, condition,
and gate together with the measures of \Cref{sec:answer_conditioned}. The answer-action ledger
(\Cref{fig:teaser}, lower left) reports the count and proportion of records in each state of
\Cref{eq:behavioral_states}, arranged as answer correctness against the action verdict so that
silent failure occupies the cell where the answer clears and the action does not. The risk map
reports \(S(t,d)\) and the record count for every task-dataset pair. The image-region audit view
(\Cref{fig:teaser}, right) presents one record at a time, showing the question, the model and
expected answers, the predicted points and ground-truth object evidence drawn on the crop, object
recall, point-in-mask precision, a coordinate-derived stray-point count, and the action verdict.

The ledger and the risk map filter a shared audit queue, the ordered set of records the audit view
steps through. Selecting a ledger state restricts the queue to that state, selecting a task--dataset
pair adds the corresponding restriction, and selecting the active pair again removes it. The two
filters combine, letting a supervisor move from a state to a particular task and dataset and then to
the individual image regions behind that pattern.

Adjusting \(\tau\) recomputes \Cref{eq:action_reliability} and everything derived from it, so the
gate governs what the audit counts as reliable rather than only what the view displays, and a
threshold-sensitivity view displays (see live dashboard) that choice as a trade between admitted risk and review effort.
Within the SFT-variant audit, changing the model condition replaces the answers and point actions on
the same image regions with \(\tau\) held fixed, and the dashboard recomputes the
audit measures across all five conditions.

\begin{figure*}[!ht]
  \centering
  \includegraphics[width=\textwidth]{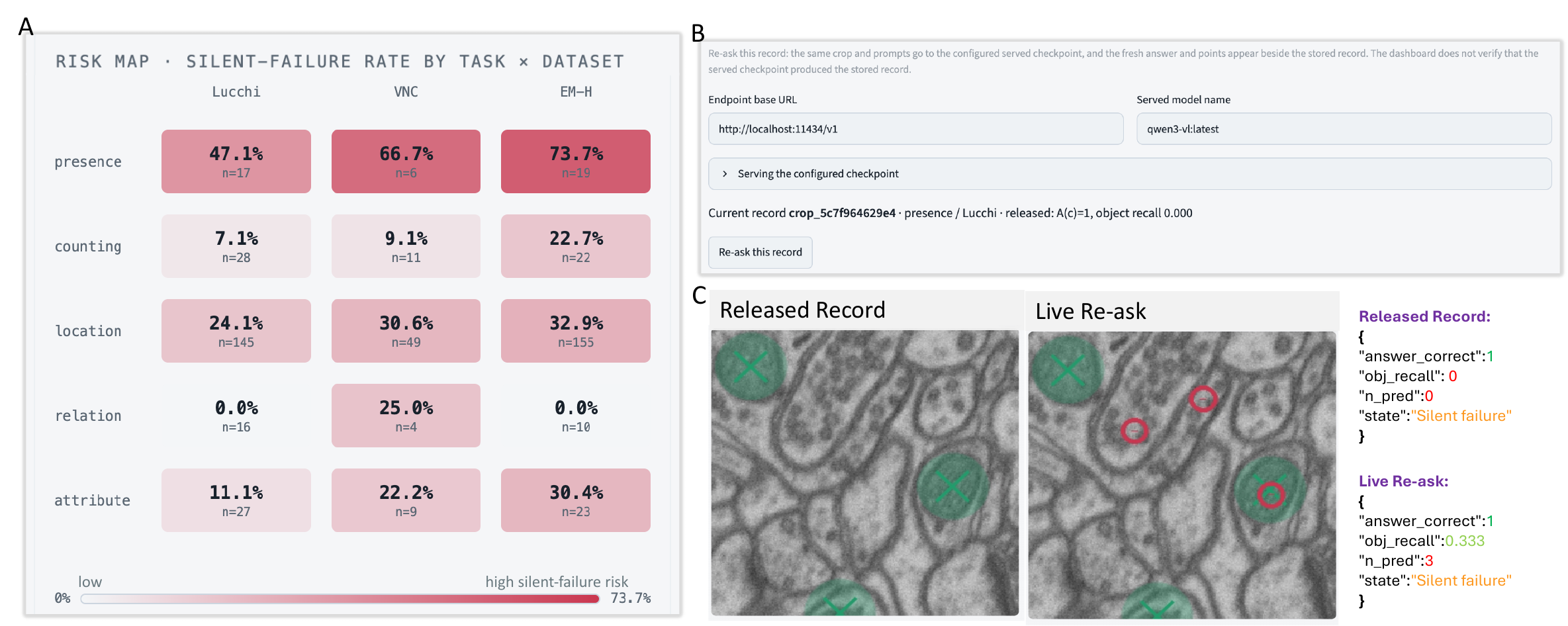}
\caption{
\textbf{Linked review beyond the run-level views.} (A) The risk map reports the silent-failure rate
and record count per task-dataset pair, and selecting a pair filters the audit queue. (B) Any
queued record can be re-asked against a served checkpoint. (C) Stored and re-asked outputs for the same region; live scoring uses the available ground-truth
coordinates. Green crosses mark centroids, red rings predicted points.
}
  \label{fig:interactions}
\end{figure*}

A supervisor can also use the dashboard to re-ask a configured served checkpoint on a record in view (\Cref{fig:interactions}B), resending both prompts and rescoring the returned outputs against the ground-truth representation available to the live dashboard.

\subsection{Human-AI Routing}
\label{sec:routing}

Once the image-level evidence has been examined, the routing panel presents the human-AI hand-off.
The supervisor records one of four dispositions, namely accepting the point action, escalating the
region for expert review, holding it for re-audit under a stricter gate, or flagging it for prompt
or model revision. Escalation and holding return the record to review, whereas flagging for model
revision accumulates a queue of image regions on which the hand-off failed, and a model revised
against that queue can be re-asked on those same records (\Cref{sec:interaction}). Each disposition is logged with the record, its answer-action state, and the value of \(\tau\) in force, so the log preserves the criterion under which each action was reviewed. These are recorded adjudications rather than
commands: the action channel is the set of normalized point coordinates parsed from the model
response, with no tool-calling or Model Context Protocol (MCP) interface~\cite{mcp2026}, and the dashboard itself
performs no segmentation, reacquisition, image-region prioritization, or microscope control.

\section{Results and Discussion}
\label{sec:results}

We trace the 541-record Qwen3-VL Staged GRPO case study end to end through the review sequence,
then recompute the same measures across model conditions on 753 matched image regions. Every
value below is reproducible in the live dashboard.

\subsection{Global Signals Reveal Weak Answer-Action Coupling}
\label{sec:global_signals}

The summary panel of \Cref{sec:interaction} reports the state of all loaded records before any
filtering, and \Cref{fig:teaser} shows it for the case study. Answer accuracy is 50.3\% over 541
image regions and mean object recall is 38.9\%, so roughly half the questions are answered
correctly while the point actions cover, on average, just over a third of the annotated
mitochondria in a region. Answer accuracy is a descriptive score for the exported questions rather
than a fixed-choice benchmark score, since the number of options \(K\) varies and 26 marked-region questions offer a single option; the Supplementary Material gives the option-count distribution and the uniform-choice
baseline of 26.3\%.

The remaining measures ask whether the two channels of the hand-off agree. At \(\tau=0.5\), silent
failures account for 27.4\% of the run, a single rate over all 541 records. Among the 272 records
whose answer matches the expected option, which is exactly the set an answer-only reviewer would
clear, 54.4\% carry a point action that fails the gate. The trust gap of \Cref{eq:trust_gap}
compares the two channels directly, at \(P(R_\tau{=}1\mid A{=}1)=45.6\%\) against
\(P(R_\tau{=}1\mid A{=}0)=39.8\%\), a difference of 5.8 percentage points whose bootstrap interval,
\([-2.9,+14.7]\), includes zero. The point-biserial (PB) correlation between answer correctness and
graded object recall is 0.061.

A correct answer is associated with only a 5.8-percentage-point higher probability of action
reliability, with an interval spanning zero, so it does not provide a reliable clearance signal in
this case study. At \(\tau=0.70\) the silent-failure rate rises to 37.9\% and the answer-conditioned failure rate to 75.4\%, while PB remains 0.061 because it uses graded object recall. The gate
changes what object coverage is accepted, not this gate-independent coupling summary.

\subsection{The Ledger Converts Failures Into Review States}
\label{sec:ledger}

A single joint score, of the kind used to combine answer and grounding quality in recent
benchmarks~\cite{nasrazadani2026vistaqa}, would compress the run into one number. The answer-action
ledger instead resolves the hand-off into the four states of \Cref{eq:behavioral_states}
(\Cref{fig:teaser}, lower left). In the case study the records divide across every state: 124
aligned passes (22.9\%), 148 silent failures (27.4\%), 107 action-only passes (19.8\%), and 162
blocked records (29.9\%).

Keeping the states apart matters because each implies a different review action. Silent failures
are the oversight hazard, since the answer clears and nothing in the language output signals a
problem, so a supervisor should prioritize these for direct inspection of the image-region
evidence. Blocked records fail on both sides and are already visible to answer-based monitoring, so
they need no separate spatial escalation. Action-only passes are the mirror case, where the point
action satisfies the gate while the answer is wrong, and they identify regions where spatial
grounding cleared the gate despite an incorrect language response. That the two outputs disagree in both
directions, on 255 of 541 records, is the argument for four categories rather than one score, which
could report that coupling is weak but not which situation a record is in. Selecting a state opens
its records in the image-region audit view, so the ledger is also the entry point to the review
path.

\subsection{The Risk Map Localizes Review Priorities}
\label{sec:riskmap}

Knowing that 27.4\% of the run consists of silent failures does not tell a supervisor where the
hand-off is fragile. The risk map therefore splits that quantity by task and dataset, reporting for
each pair the share of its records that are silent failures together with the record count behind
the share (\Cref{fig:interactions}A). Presence carries the highest shares in every dataset, at
47.1\% for Lucchi, 66.7\% for VNC, and 73.7\% for EM-H, so correct yes-or-no answers about whether
mitochondria are present are frequently paired with point actions that miss them. Location, which
groups location and marked-region questions, shows consistent risk between 24.1\% and 32.9\%.

The pair with the highest share is not the pair holding the most silent failures, which is why the
map reports both quantities. Presence carries the highest shares but rests on 42 records in total,
whereas the location row contributes 101 of the 148 silent failures in the run. A supervisor
reading color alone would prioritize presence, while one reading the counts alongside it would
spend most of the available effort on location. The map thus converts the single 27.4\%
silent-failure rate into localized review priorities, and high-value pairs can be sent for
image-region inspection, re-audited under a stricter gate, or used to target prompt and model
revision. Pairs with fewer than 20 records are therefore prompts to inspect the underlying examples
rather than stable estimates of a task or dataset weakness.

\subsection{The Audit Path Traces Both Hand-offs to a Recorded Decision}
\label{sec:audit_path_sec}

\Cref{fig:teaser} shows one case-study record at the end of the review path. Selecting the
silent-failure state opens its 148 records in the image-region audit view. For the record shown the answer is scored correct, so an answer-only reviewer would clear the region, while the paired action emits three points that all miss the annotated mitochondria,
giving object recall 0.00 and a failing verdict. The dashboard's coordinate-derived diagnostic
also flags all three emitted points as stray. Nothing in the language output indicates this.
It becomes visible only because the answer, the proposed points, and the annotated objects are shown
against the same image. That joint inspection is how the dashboard audits the language-to-action
hand-off.

Point-level errors are not confined to failing records. Mean point-in-mask precision among the 272
answer-correct records is 0.471, and of the 231 records whose action passes the gate, 172 are also
flagged with at least one stray point by the coordinate-derived diagnostic. Passing the gate
therefore does not exhaust what a reviewer needs to see, which is why the review path ends at image
evidence rather than at a threshold.

The routing panel beneath the audit view presents the human-AI hand-off. The evidence here argues
against accepting the action, and whichever disposition the supervisor selects is logged with the
answer-action state and the gate it was taken under (\Cref{sec:routing}), so the log preserves the criterion under which each action was reviewed.

\subsection{Re-audit Across Model Conditions}
\label{sec:sft_variant_audit}

\begin{table}[t]
\centering
\caption{
Re-audit across model conditions on 753 matched image regions at \(\tau=0.5\). Answer accuracy,
mean object recall, and joint silent-failure rate are percentages. Trust gap is in percentage points with
its bootstrap interval; PB is the point-biserial correlation.
}
\label{tab:qwen_conditions}
\resizebox{\columnwidth}{!}{
\begin{tabular}{lccccc}
\hline
Condition & Answer & Recall & Silent & Trust gap & PB \\
\hline
Zero-shot & 43.7 & 21.4 & 33.6 & +2.6 [-3.2, +8.6] & 0.023 \\
Perception SFT & 50.6 & 25.6 & 38.1 & -2.5 [-8.7, +4.1] & -0.009 \\
Grounding SFT & 43.2 & 23.1 & 31.6 & +3.6 [-2.6, +10.3] & 0.033 \\
Staged SFT & 50.6 & 37.1 & 29.7 & +3.6 [-3.4, +10.8] & 0.043 \\
Joint SFT & 52.9 & 32.5 & 33.5 & +2.6 [-4.3, +9.4] & 0.028 \\
\hline
\end{tabular}
}
\end{table}

Switching the audit source to the matched 753-region source recomputes every measure above for
Zero-shot and the four supervised variants of \Cref{sec:setup}, on identical image regions and at
the same gate (\Cref{tab:qwen_conditions}). All five share the same question set, so the comparison is unaffected
by question composition. Answer accuracy ranges from 43.2\% to 52.9\% and mean object
recall from 21.4\% to 37.1\%, and the two do not move together.

Because a silent failure requires a correct answer, improving answer accuracy without a
corresponding improvement in action reliability can increase rather than reduce the hazard. Zero-shot and Perception SFT illustrate this pattern, since answer accuracy rises from 43.7\% to 50.6\% and the silent-failure rate rises with it, from 33.6\% to 38.1\%. Perception
SFT and Staged SFT are sharper still, reporting identical answer accuracy of 50.6\% while their
silent-failure rates differ by 8.4 percentage points, 38.1\% against 29.7\%, tracking an
11.5-percentage-point difference in mean object recall. A supervisor choosing between the two on
answer accuracy would see no difference where the audit sees a substantial one.

Furthermore, answer-action coupling remains weak across all five conditions. Every trust gap lies
between \(-2.5\) and \(+3.6\) percentage points, every interval includes zero, and PB stays between
\(-0.009\) and 0.043. Joint SFT reaches the highest answer accuracy while its interval still spans
zero, and Zero-shot shows similarly weak coupling, so weak coupling is not confined to conditions
that supervise the two channels separately.

The audit can also be repeated for a single record. \Cref{fig:interactions}B--C shows a presence
record from Lucchi whose stored action returned no points at all, and re-asks it against a
separately served checkpoint. The re-asked action emits three points,  one of which matches a ground-truth centroid,
giving live object recall 0.333. Both answers are correct and neither action
clears the gate, so the region is a silent failure under both. Switching condition and
re-asking are therefore the two scales at which the audit repeats, and both let a supervisor revisit failures at aggregate and
record levels rather than only through an aggregate score.

\section{Conclusion and Future Work}
\label{sec:conclusion}

We presented \emph{Spatial Action Review}, an interactive visual analytics dashboard for auditing
language-to-action hand-offs in EM image analysis. It carries a supervisor from answer-action
states and task-dataset risk to the image-region evidence behind them, allowing a proposed point
action to be checked against the annotated objects it must cover before downstream use. The review
ends in the human-AI hand-off, where the disposition is logged with the active gate and
answer-action state.

Using an EM-adapted Qwen3-VL Staged GRPO case study, we demonstrate why this visual oversight gap matters.
Correct answers were often paired with unreliable point actions, producing a 27.4\% silent-failure
rate, while answer correctness provided little separation between reliable and unreliable actions.
Repeating the audit across five model conditions likewise finds weak answer-action coupling,
including when answer accuracy improves. Better answer accuracy therefore does not by itself
provide evidence that the paired action is more reliable.

The broader lesson is that human-readable answers should not be treated as sufficient clearance for
spatial outputs consumed downstream. Although demonstrated in EM mitochondria analysis, the
paired-record schema and review path apply wherever a model emits an answer and a spatial output
for the same image. Future work should extend the audit to richer spatial outputs, including object
masks and candidate fields of view, measure downstream and reviewer effects, and connect reviewed
actions to scientific tools through explicit tool-calling interfaces (e.g., MCP) while preserving
human oversight.

% \section*{Supplemental Materials}
% \label{sec:supplemental_materials}

\acknowledgments{
This project was funded by a Wellcome Trust Investigator Award (Ref: 205038/Z/16/Z), an ERC grant (Ref: ERC-2018-COG: 819650), a BBSRC Grant (Ref: APP26929, ``Population connectomics") and MRC LMB core funding.
}

\bibliographystyle{abbrv-doi}

\bibliography{references}
\clearpage
% Supplementary material for the VGTC conference style template.
% Compile from a VGTC LaTeX project that contains vgtc.cls and the VGTC bibliography style files.
% ============================================================
% Supplementary Material
% ============================================================

\onecolumn

% Start supplementary numbering
\setcounter{section}{0}
\setcounter{figure}{0}
\setcounter{table}{0}
\setcounter{equation}{0}

\renewcommand{\thesection}{S\arabic{section}}
\renewcommand{\thefigure}{S\arabic{figure}}
\renewcommand{\thetable}{S\arabic{table}}
\renewcommand{\theequation}{S\arabic{equation}}

\section*{Supplementary Material: Spatial Action Review}

\begin{center}
\begin{minipage}{1\textwidth}
  \centering

  \includegraphics[
    width=\linewidth,
    height=0.88\textheight,
    keepaspectratio
  ]{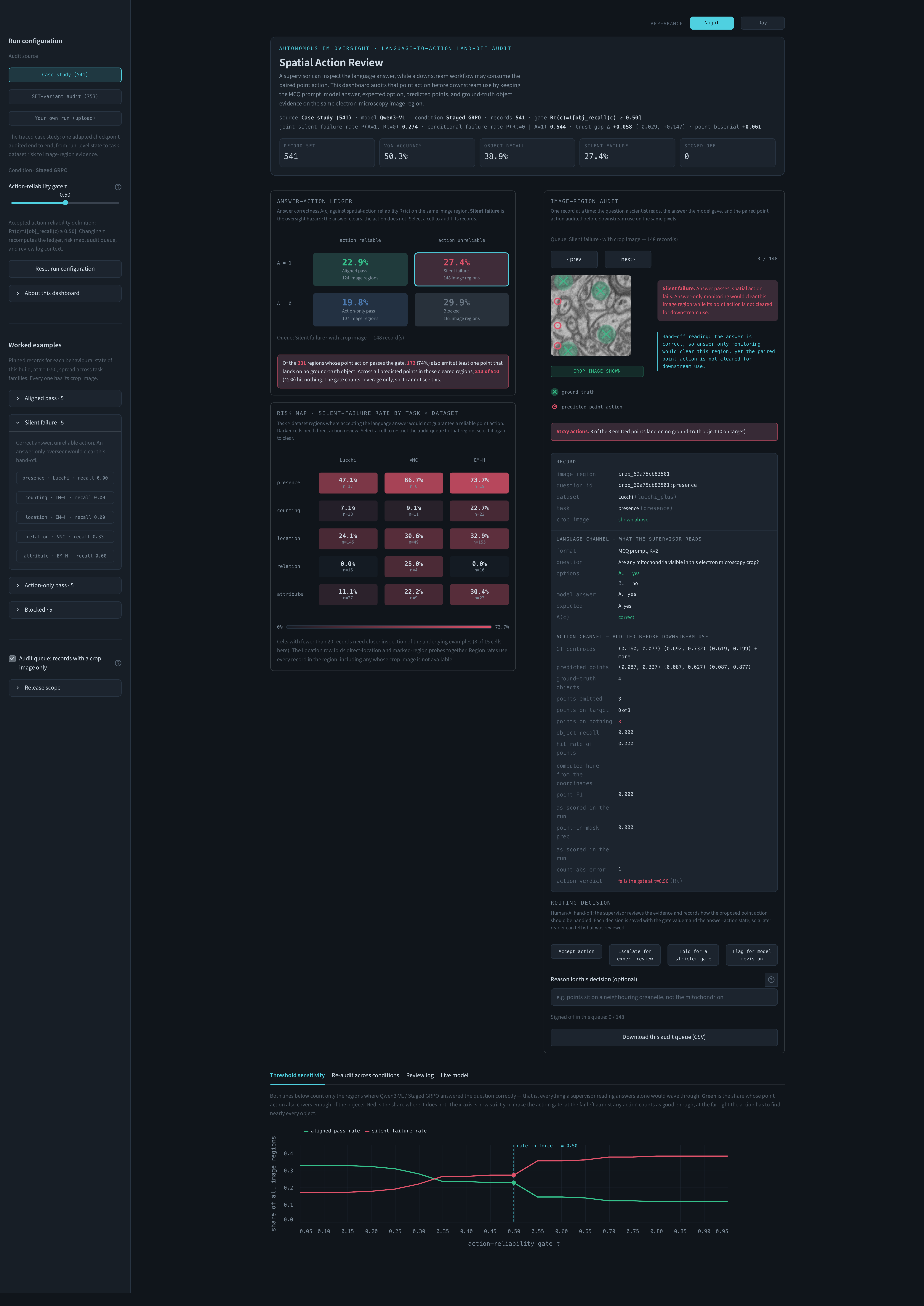}

  \captionof{figure}{
  \textbf{Dark-mode view of the interactive \emph{Spatial Action Review} dashboard}
  for the 541-record case study at \(\tau=0.5\). The sidebar selects the audit
  source, the gate, and the model condition. The summary panel and the
  answer-action ledger report the run, the risk map locates silent-failure
  risk across task and dataset, and the image-region audit opens one record,
  where the answer is correct while no predicted point falls inside an
  annotated mitochondrion, giving object recall \(0.00\) and \(R_\tau=0\).
  The record detail panel below it separates the language channel a supervisor
  reads from the action channel audited before downstream use, and the routing
  panel records one of four dispositions. At the foot, the threshold-sensitivity
  view plots the aligned-pass and silent-failure rates across \(\tau\).
  }
  \label{fig:supp_interactive_dashboard}

\end{minipage}
\end{center}

\clearpage
\twocolumn

\section{Supplement Scope}
\label{sec:supp_scope}

This supplement documents the artifacts needed to interpret the paper, namely the interactive dashboard view, the paired answer-action record, the composition of the two reported audit sources, the scoring conventions behind the reported measures, and the interaction details of the dashboard.

\section{Paired Answer-Action Records}
\label{sec:supp_records}

Each dashboard entry corresponds to one electron microscopy image region. The record links the
language output that a supervisor can inspect with the point action a downstream stage may consume.
Table~\ref{tab:supp_record_schema} summarizes the fields used by the dashboard.

Each image region is a \(256\times256\)-pixel two-dimensional grayscale crop taken from the
annotated EM volume with stride 128. The same crop and labeling conventions are used to construct
the ground-truth object evidence against which point actions are scored.

\begin{table}[H]
\centering
\small
\caption{Fields in the paired answer-action record used by the dashboard. The same record supports
the ledger, risk map, and image-region audit view.}
\label{tab:supp_record_schema}
\begin{tabularx}{\columnwidth}{>{\raggedright\arraybackslash}p{0.34\columnwidth}X}
\toprule
Field group & Role in the dashboard \\
\midrule
Image-region identity & Connects the audit record to the source dataset and image region. \\
Task metadata & Stores the normalized task type used for grouping records in the risk map. \\
VQA channel & Stores the multiple-choice question, answer options, expected answer, and model answer. \\
Point-action channel & Stores the predicted normalized point set returned for the image region. \\
Object evidence & Stores ground-truth mitochondria evidence used to score spatial reliability. \\
Spatial metrics & Stores object recall and other diagnostic quantities used in the audit view. \\
Review state & Stores the derived answer-action state used by the ledger. \\
\bottomrule
\end{tabularx}
\end{table}

The VQA channel is a multiple-choice answer channel. The point-action channel is an all-object
spatial channel: the model is asked to return one normalized point for each visible mitochondrion
in the image region. The action gate therefore measures coverage of labeled objects rather than
whether a single witness point is present. In the dashboard, object recall is converted into a
binary action-reliability indicator,
\begin{equation}
R_\tau(c)=\mathbbm{1}[\mathrm{obj\_recall}(c)\geq\tau],
\end{equation}
where \(c\) denotes an image region and \(\tau\) is the action-reliability gate. The primary reported
analyses use \(\tau=0.5\), while the dashboard allows the supervisor to vary the gate during
re-audit. The graded object-recall value remains visible in the image-region audit panel.

The paired scores define four review states. Aligned pass records have a correct answer and a
reliable action. Silent failures have a correct answer and an unreliable action. Action-only
passes have an incorrect answer and a reliable action. Blocked records have an incorrect answer
and an unreliable action. This state decomposition is the basis for the answer-action ledger in
the dashboard.

\section{Audit-Set Composition}
\label{sec:supp_composition}

The matched audit source contains 753 image regions, each evaluated under the same five model
conditions reported in the paper. For any selected condition, these form 753 paired
image-region records. The set includes image regions from Lucchi, EM-H, and VNC.
Table~\ref{tab:supp_dataset_distribution} reports the image-region distribution, and
Table~\ref{tab:supp_task_distribution} reports the normalized task distribution.

\begin{table}[h]
\centering
\small
\caption{Image-region distribution in the two reported audit sources.}
\label{tab:supp_dataset_distribution}
\begin{tabular}{lcc}
\toprule
Dataset & Case study (541) & Matched (753) \\
\midrule
Lucchi & 233 & 342 \\
EM-H   & 229 & 332 \\
VNC    &  79 &  79 \\
\bottomrule
\end{tabular}
\end{table}

\begin{table}[h]
\centering
\small
\caption{Task distribution in the two reported audit sources. Marked-region probes are grouped
with location probes in the normalized task field.}
\label{tab:supp_task_distribution}
\begin{tabular}{lcc}
\toprule
Task & Case study (541) & Matched (753) \\
\midrule
Attribute & 59 & 79 \\
Counting  & 61 & 80 \\
Location, including marked-region probes & 349 & 486 \\
Presence  & 42 & 68 \\
Relation  & 30 & 40 \\
\bottomrule
\end{tabular}
\end{table}

The multiple-choice probes have a variable number of answer options.
Table~\ref{tab:supp_option_distribution} reports the option-count distribution for both sources,
together with the record-wise uniform-choice baseline. No probe uses seven or eight options.

\begin{table}[h]
\centering
\small
\caption{Multiple-choice option-count distribution in the two reported audit sources, with the
record-wise uniform-choice baseline \(\mathrm{mean}(1/K)\).}
\label{tab:supp_option_distribution}
\begin{tabular}{lcc}
\toprule
Number of options \(K\) & Case study (541) & Matched (753) \\
\midrule
1 &  26 &  32 \\
2 &  69 & 105 \\
3 &  19 &  30 \\
4 & 113 & 156 \\
5 & 103 & 144 \\
6 &  61 &  80 \\
9 & 150 & 206 \\
\midrule
Uniform-choice baseline & 0.263 & 0.264 \\
\bottomrule
\end{tabular}
\end{table}

\section{Marked-Region Probes}
\label{sec:supp_marked_regions}

Marked-region probes ask the model to choose which marked image region contains a mitochondrion. The regions are represented by normalized bounding boxes in the question text, and the answer options are the region labels. Some marked-region probes contain a single candidate region and therefore have \(K=1\). These records are included in the reported answer-correctness and reliability summaries. Their answer-side accuracy should be read as a descriptive score rather than as fixed-choice benchmark accuracy when \(K=1\).

In the task-by-dataset risk map, marked-region probes are grouped with location probes. This grouping keeps the risk map compact and reflects the shared visual task: both probe families require resolving spatial evidence in the image region. The point-action channel remains the same all-object point request for every task type.

\section{Relationship Between the Two Audit Sources}
\label{sec:supp_exports}

The paper reports results from two audit sources because they serve different roles in the analysis. The
541-record source is the dashboard case study: one adapted checkpoint, Staged GRPO, traced end to
end from run-level summaries to task-dataset risk to image-region evidence
(Figure~\ref{fig:supp_interactive_dashboard}). The matched audit source contains 753 image regions
and is used to compare answer-action reliability across five model conditions under a shared
record schema and scoring rule.

The two sources share the same record definition and action-reliability gate, and their image
regions are drawn from the same pool: all 541 case-study identifiers occur among the 753, with 212
further identifiers exercised only in the cross-condition comparison.
Table~\ref{tab:supp_export_origin} summarizes the relationship.

\begin{table}[h]
\centering
\small
\caption{Relationship between the dashboard case-study and the matched audit sources.}
\label{tab:supp_export_origin}
\begin{tabularx}{\columnwidth}{>{\raggedright\arraybackslash}p{0.46\columnwidth}cc}
\toprule
Quantity & Dashboard & Matched \\
\midrule
Image-region identifiers & 541 & 753 \\
Identifiers shared with the other source & 541 & 541 \\
Identifiers unique to source & 0 & 212 \\
Primary role & Case study & Re-audit \\
\bottomrule
\end{tabularx}
\end{table}

\section{Spatial Reliability and Diagnostic Fields}
\label{sec:supp_diagnostics}

Object recall is the primary spatial score used by the dashboard because the point-action channel
asks for all visible mitochondria in the image region. The binary action gate \(R_\tau\) is derived
from this score and drives the answer-action ledger, silent-failure rate, trust gap, and risk map.

The image-region audit view also reports additional spatial diagnostics. Point F1 balances missed
objects against extra predicted points, point-in-mask precision measures the fraction of emitted
points that fall inside an annotated object mask, and count absolute error is the difference
between the predicted point count and the labeled object count. The coordinate-derived hit rate is
described in Section~\ref{sec:supp_scoring}. These diagnostics support image-region review but do
not change the binary ledger state used in the main dashboard analysis.

\section{Scoring Conventions}
\label{sec:supp_scoring}

Six case-study image regions contain no annotated mitochondria. Object recall is defined as one
when the model returns no points and zero otherwise, so a region correctly left empty counts as
reliable while one receiving spurious points does not. All six contained at least one predicted point and are therefore scored unreliable.

Object recall and point-in-mask precision are both computed against the annotation label image,
testing whether a predicted point falls inside an object mask. Every record in the two reported
sources is scored on this mask-based path. Where object masks are unavailable but centroids are
supplied, object coverage can instead be evaluated by assigning predicted points to centroids
under a normalized distance threshold of 0.1; neither reported source requires this fallback.
Point-in-mask precision requires an annotation label image.

The dashboard additionally derives a point hit rate from the coordinates alone, using one-to-one
assignment between predicted points and object centroids under the same normalized distance
threshold of 0.1. This diagnostic can therefore be computed whenever an export supplies predicted
points and object centroids, whereas point-in-mask precision requires the annotation label image
and is read from the scored record. The dashboard labels ``points on target,'' ``points on
nothing,'' and ``stray actions'' refer to this coordinate-derived diagnostic rather than to
point-in-mask membership. The two quantities use different notions of a successful point.
Point-in-mask precision tests whether each emitted point lies inside an annotated object mask,
whereas the hit rate tests whether it can be matched to a distinct object centroid within the
distance threshold. A point may therefore be inside an object mask but too far from its centroid
to count as a hit, while a point sufficiently close to a centroid is not guaranteed by the
distance rule alone to lie inside that object's mask.

The one-to-one assignment also treats repeated predictions differently. If multiple predicted
points fall on the same annotated object, each can contribute to point-in-mask precision, whereas
at most one can be assigned to that object's centroid for the hit-rate calculation. Consequently,
neither point-in-mask precision nor the coordinate-derived hit rate is guaranteed to be larger
than the other. Their difference reflects both the distinction between mask membership and
centroid proximity and the one-to-one assignment used by the coordinate-derived diagnostic.

\section{Interaction Details}
\label{sec:supp_interaction}

The threshold-sensitivity view plots the aligned-pass and silent-failure rates across the available
range of \(\tau\) and tabulates, for each value, how many actions the gate would clear and how many
it would hold for review (see Figure~\ref{fig:supp_interactive_dashboard}).
Because the two rates sum to answer accuracy at every gate, the view
separates what the gate can change from what it cannot. Moving the gate does not alter how many
answers are correct, only how many correct answers rest on an action that covers enough of the
region.

The re-ask connection uses an OpenAI-compatible chat-completions endpoint and stays disabled until
a base URL and a served model name are supplied. The re-ask is sent to the explicitly configured
served checkpoint; the dashboard does not verify that this checkpoint produced the stored record.
A re-ask leaves the stored record and every aggregate view unchanged.

Each routing disposition is logged with the record and crop identifiers, the audit source, model
condition, task, dataset, answer-action state, the value of \(\tau\) in force, object recall,
answer correctness, the selected disposition, an optional written reason, and a timestamp. A later
change of gate does not rewrite the gate a decision was taken under. Decisions are held in the
active session and can be exported as CSV or as a self-describing JSON document carrying the audit
context, the disposition vocabulary, and per-disposition totals.

\end{document}